\documentclass[sn-mathphys,Numbered]{sn-jnl}
\usepackage{graphicx}%
\usepackage{multirow}%
\usepackage{amsmath,amssymb,amsfonts}%
\usepackage{amsthm}%
\usepackage{mathrsfs}%
\usepackage[title]{appendix}%
\usepackage{xcolor}%
\usepackage{textcomp}%
\usepackage{manyfoot}%
\usepackage{booktabs}%
\usepackage{algorithm}%
\usepackage{algorithmicx}%
\usepackage{algpseudocode}%
\usepackage{listings}%
\usepackage{tabularx}
\usepackage[inkscapearea=page]{svg}
\usepackage{adjustbox}
\usepackage{caption}
\usepackage{subcaption}
\usepackage{lineno}

\theoremstyle{thmstyleone}%

\theoremstyle{thmstyletwo}%

\theoremstyle{thmstylethree}%

\begin{document}
\title[OCT image classification with Federated Learning]{Investigation of Federated Learning Algorithms for Retinal Optical Coherence Tomography Image Classification with Statistical Heterogeneity}

\author*[1, 2]{\fnm{Sanskar} \sur{Amgain}}\email{sanskar.amgain@naamii.org.np}
\equalcont{These authors contributed equally to this work.}

\author[1, 2]{\fnm{Prashant} \sur{Shrestha}}
\equalcont{These authors contributed equally to this work.}
\author[2, 4]{\fnm{Sophia} \sur{Bano}} 
\author[2]{\fnm{Ignacio} del Valle \sur{Torres}} 
\author[2]{\fnm{Michael} \sur{Cunniffe} }
\author[2]{\fnm{Victor} \sur{Hernandez}} 
\author[2]{\fnm{Phil} \sur{Beales}}
\author[2, 3]{\fnm{Binod} \sur{Bhattarai}}

\affil[1]{\orgname{NAAMII}, \orgaddress{\city{Lalitpur}, \country{Nepal}}}
\affil[2]{\orgname{Base Gene Therapeutics Limited}, \orgaddress{\country{UK}}}
\affil[3]{\orgname{University of Aberdeen, UK}}
\affil[4]{\orgname{WEISS, and  University College London, UK}}

\newcommand{\prashant}[1]{\textcolor{purple}{PS: #1}}
\newcommand{\sanskar}[1]{\textcolor{blue}{SA: #1}}

\abstract{
\par

\textbf{Purpose:} We apply federated learning to train an OCT image classifier simulating a realistic scenario with multiple clients and statistical heterogeneous data distribution where data in the clients lack samples of some categories entirely. 
\par
\textbf{Methods:} We investigate the effectiveness of FedAvg and FedProx to train an OCT image classification model in a decentralized fashion, addressing privacy concerns associated with centralizing data. We partitioned a publicly available OCT dataset across multiple clients under IID and Non-IID settings and conducted local training on the subsets for each client. We evaluated two federated learning methods, FedAvg and FedProx for these settings.
\par
\textbf{Results:} 
Our experiments on the dataset suggest that under IID settings, both methods perform on par with training on a central data pool. However, the performance of both algorithms declines as we increase the statistical heterogeneity across the client data, while FedProx consistently performs better than FedAvg in the increased heterogeneity settings. 
\par
\textbf{Conclusion:} 
Despite the effectiveness of federated learning in the utilization of private data across multiple medical institutions, the large number of clients and heterogeneous distribution of labels deteriorate the performance of both algorithms. Notably, FedProx appears to be more robust to the increased heterogeneity.
}


\maketitle

\section{Introduction}\label{introduction}
\par
Optical Coherence Tomography~(OCT) is a widely used imaging technique for diagnosing retinal abnormalities. An OCT image captures the different cross-sectional layers of the retina, the thickness of which is important for performing diagnosis~\cite{al2013optical}. A large number of works have explored the use of artificial intelligence-based methods for assistive and automated diagnosis from OCT images. The performance and generalizability of these models are directly correlated with the amount of data available for training. However, due to the privacy concerns that arise with sharing raw medical data outside an institution, multi-institution collaboration for the curation of a large central database remains a challenging problem. 

Few works have recently explored the use of federated learning for OCT image classification~\cite{gholami2023federated, ran2023developing} bypassing the need for a central data curation.
For instance, Gholami et al.~\cite{gholami2023federated} used federated learning to classify OCT images into normal and age-related macular degeneration (AMD) categories while Ran et al.~\cite{ran2023developing} uses FedAvg to detect glaucoma samples. 
However, these works \emph{do not consider the statistical heterogeneous setting} where the local data in certain clients lack instances of some labels entirely. Furthermore, these works experiment with a  \emph{small number of clients (3-4)} sidestepping the potential issues that arise when dealing with many clients such as the increased communication overhead. Our work studies two prominent federated learning methods on OCT images simulating such label heterogeneity with a larger number of clients. 

\section{Methods}\label{methods}
\begin{figure}[ht]
    \centering
    \includegraphics[width=\linewidth]{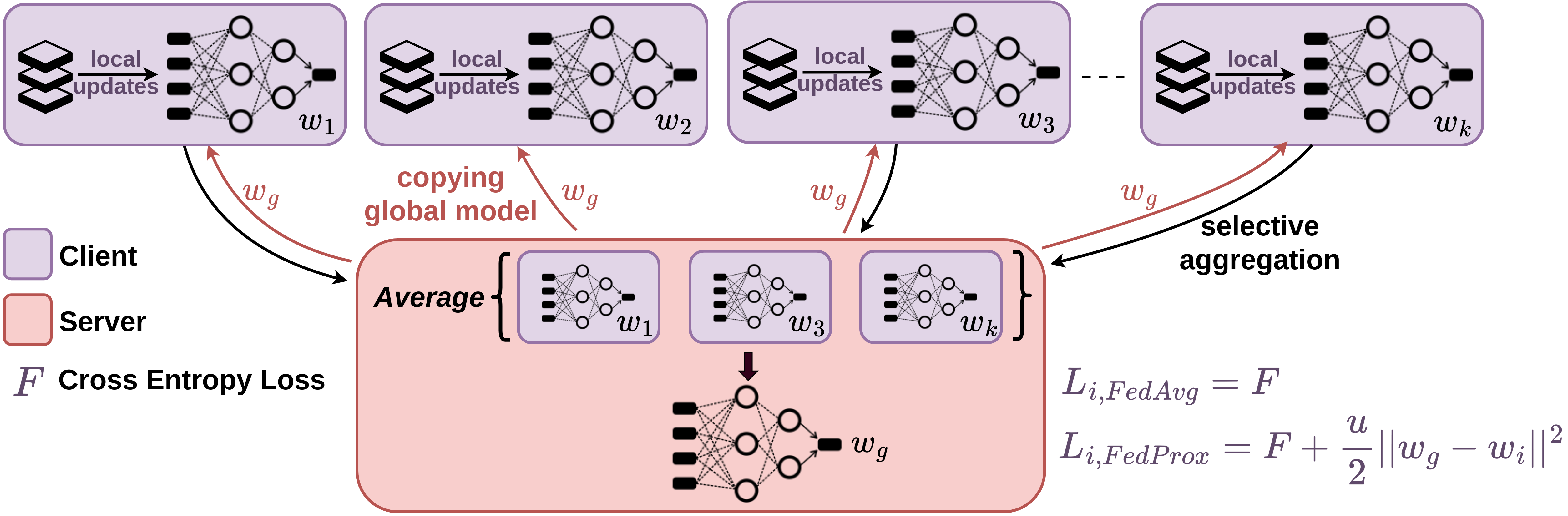}
    \caption{Illustration of FedAvg and FedProx training process. In each communication round, a subset of clients trained locally on their respective data splits with objective $L_{method}$ are selected for aggregation and used to update the global model. The updated global model parameters, $w_g$ are copied back to all the clients.}
    \label{fig:fedavg}
\end{figure}
We divided the OCT dataset into disjoint subsets, each associated with a separate client; simulating separate private data associated with distinct medical institutions. 
As shown in Fig. ~\ref{fig:fedavg}, in each communication round, clients initialize their local parameters with the current global model parameters $w_g$. Each client, $i$ is trained on their respective data split with objective $L_{i, method}$, the clients send back updated model parameters $\{w_1, w_2, \ldots w_k\}$ to the server, which averages the parameters across clients to obtain the updated global model. This process is repeated until convergence. Since the sharing of information during the training process occurs only at the level of model parameters, and not raw medical data, federated learning achieves the aggregation of information from multiple clients/institutions in a privacy-preserving manner. 

To reduce communication overhead, a random subset of the clients is selected for local training and aggregation. We experiment with two federated learning methods: FedAvg~\cite{mcmahan2017communication} and FedProx~\cite{li2020federated}. FedAvg employs standard cross-entropy loss, $F$ as the local training objective $L$, while FedProx additionally penalizes the deviation of local parameters from the global model. Our study utilized the OCT images from Kermany et al.~\cite{kermany2018identifying} comprising 84,484 OCT images categorized as CNV, DRUSEN, DME, or NORMAL (Fig.~\ref{fig:samples}).

\section{Experiments}\label{Experiments}

\begin{figure}[h!]
\begin{subfigure}[b]{0.69\linewidth}
    \centering
    \includegraphics[width=\textwidth]{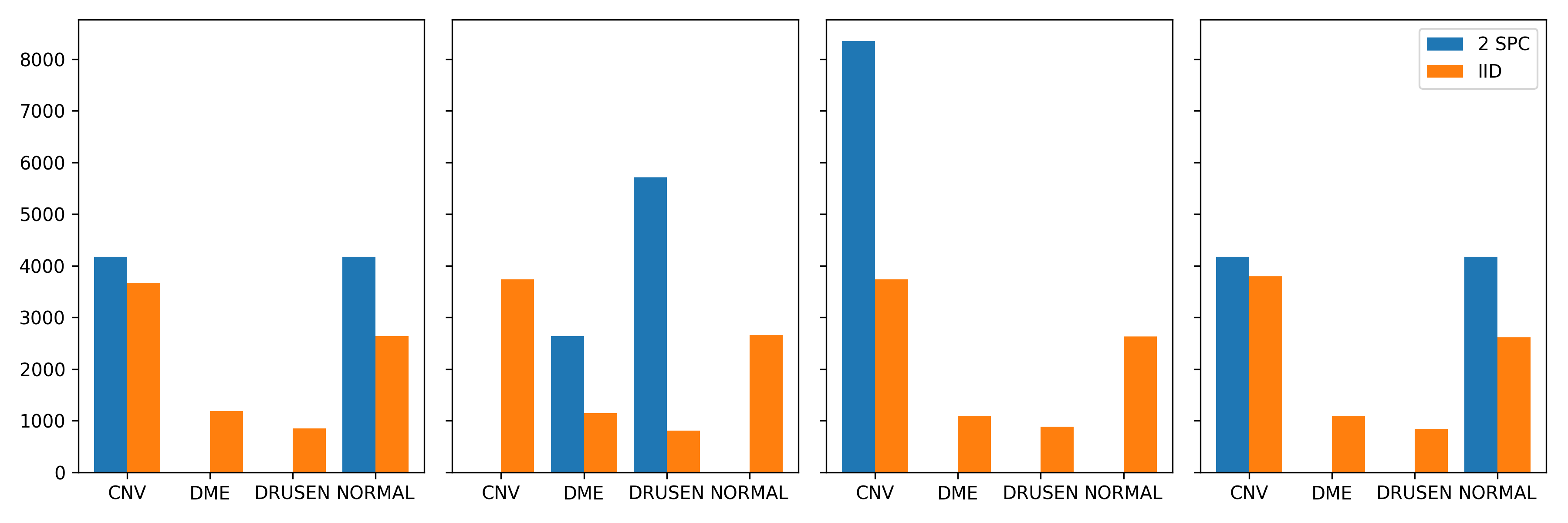}
    \caption{}
    \label{fig:label_distribution}
\end{subfigure}
\hfill
\begin{subfigure}[b]{0.30\linewidth}
    \centering
    \includegraphics[width=\textwidth]{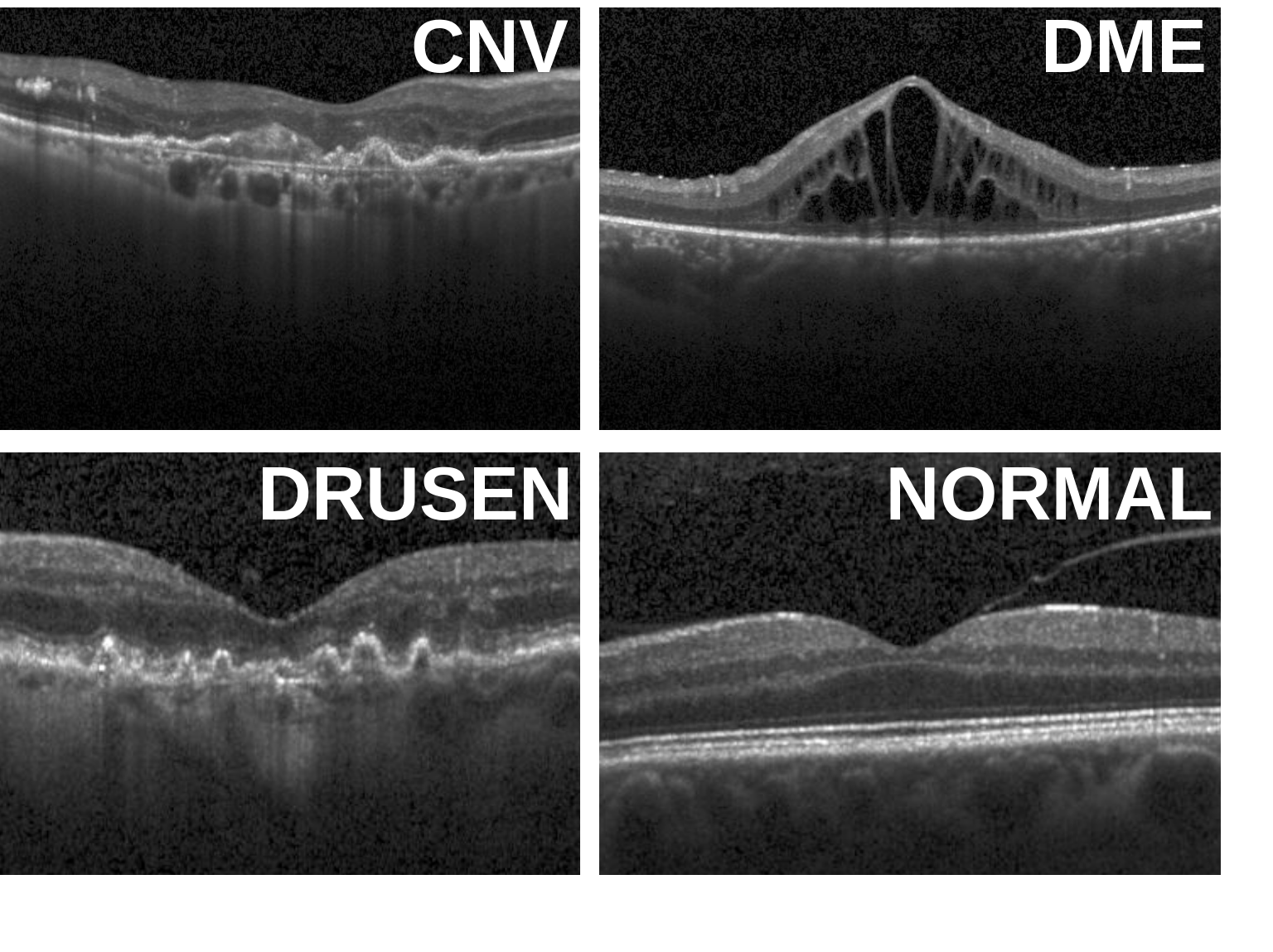}
    \caption{}
    \label{fig:samples}
\end{subfigure}
\caption{(a) Label distribution of 4 randomly selected clients for IID and 2 SPC Non-IID settings. For the IID setting, the distribution of labels across clients is similar, while for 2 SPC, every client has multiple missing labels and distinct distributions (b) Image samples from the dataset}
\end{figure}

We evaluated the effectiveness of two prominent federated learning methods, FedAvg and FedProx for the OCT image classification task. With a total of 10 clients, we randomly selected 50\% for aggregation to minimize communication overhead. We experimented with two data configurations: IID and Non-IID splits. 
For the IID setting, we randomly partitioned the data into 10 disjoint subsets, each associated with a separate client. 
The data splits are thus less likely to have missing labels and exhibit less heterogeneity in label distribution across clients. 
For the Non-IID setting, we grouped the images by labels and divided them into $10*k$ shards, each containing images belonging to a single category. Each client is then assigned $k$ shards as the local data. When $k < n$, $n$ being the total number of categories, this sharding guarantees that no client possesses samples from all the categories, mirroring real-life statistical heterogeneity where \emph{not all medical centers have the same distribution of patient data, and access to all the categories}. Fig~\ref{fig:label_distribution} illustrates the label distribution for a random subset of clients in both IID and Non-IID settings.

For our classification model, we initialized the encoder with ResNet18 parameters pretrained on ImageNet. The backbone of the network is frozen and only the classification head is fine-tuned on the OCT images. We use SGD optimizer with a learning rate of 0.01 and batch size of 64 to train each client model for 2 local epochs per communication round. The training is performed for 100 communication rounds. We use average accuracy to evaluate the performance of the models under IID and $k \in \{2, 3\}$ Non-IID settings.

\section{Results}\label{Results}
As shown in Table~\ref{tab:results}, both FedAvg and FedProx perform well under IID settings scoring on par with 98.933$\pm$0.27 obtained with centralized training.
However as the statistical heterogeneity across the clients is increased, under the 3 shards per client (2SPC) and 2 shards per client (2SPC) Non-IID settings, the performance of both approaches begins to deteriorate. 
Notably, FedProx consistently outperforms FedAvg in our Non-IID settings. This is attributed to the proximal loss term of FedProx that constrains the local updates to be closer to the global model. This constraint mitigates the effect of statistical heterogeneity leading to better performance. 
\begin{table}[h]
    \centering
    \begin{tabular}{|c|c|c|c|}
        \hline
         Method &  IID & 2SPC & 3SPC\\
         \hline
        ~FedAvg~ & \textbf{~98.93 $\pm$ 0.05~} & ~87.22 $\pm$ 5.97~ & ~88.32 $\pm$ 2.30~ \\
        \hline
        ~FedProx ($\mu = 0.2$)~ & ~98.80 $\pm$ 0.10~ & \textbf{~90.57 $\pm$ 3.09~} & \textbf{~91.01 $\pm$ 2.51~} \\
        \hline
    \end{tabular}
    \caption{Accuracy on test dataset under IID, 2SPC, 3SPC data splits. Results consist of mean and standard deviation over 3 random seeds.}
    \label{tab:results}
\end{table}

\section{Conclusion}\label{conclusion}
Federated learning provides an effective solution to utilize private data across multiple medical institutions, achieving high performance without exposing raw medical data.  
Our study investigated the effectiveness of two federated learning methods, FedAvg and FedProx under different heterogeneity settings for OCT image classification. Both methods consistently performed well across diverse scenarios. However, as the statistical heterogeneity across clients increases, the performance of both approaches starts to deteriorate. Notably, FedProx demonstrated greater robustness to the increased heterogeneity than FedAvg. 
This suggests that in highly heterogeneous settings, the application of federated learning for OCT image classification necessitates more sophisticated approaches.
Beyond statistical heterogenity, future work can investigate the effectiveness of federated learning methods with separate domain specific medical data in each client, along with dynamic and evolving medical datasets.

\backmatter

\bigskip

\begin{appendices}

\end{appendices}

\bibliography{sn-bibliography}

\end{document}